\documentclass[conference]{IEEEtran}
\IEEEoverridecommandlockouts
\usepackage{amsmath,amssymb,amsfonts}
\usepackage{algorithm}
\usepackage{algorithmicx}
\usepackage{algpseudocode}
\usepackage{graphicx}
\usepackage{textcomp}
\usepackage{xcolor}
\usepackage{hyperref}
\usepackage{array}
\usepackage{stfloats}
\usepackage{url}
\usepackage{subfig}
\usepackage{cite}
\usepackage{multirow}
\usepackage{multirow}
\usepackage{booktabs} 
\usepackage{makecell}
\usepackage{times}
\usepackage[T1]{fontenc}

\def\BibTeX{{\rm B\kern-.05em{\sc i\kern-.025em b}\kern-.08em
    T\kern-.1667em\lower.7ex\hbox{E}\kern-.125emX}}
\begin{document}

\title{DynaSplat: Dynamic-Static Gaussian Splatting with Hierarchical Motion Decomposition for Scene Reconstruction}

% \author{Anonymous ICME submission}

\author{
Junli Deng\textsuperscript{1}, 
Ping Shi\textsuperscript{1}\textsuperscript{$\dagger$}, 
Qipei Li\textsuperscript{1}, 
Jinyang Guo\textsuperscript{2} \\
\textsuperscript{1}Communication University of China, Beijing, China \\
\textsuperscript{2}Beihang University, Beijing, China \\
\{dengjunliok, shiping, liqipei666\}@cuc.edu.cn, 
jinyangguo@buaa.edu.cn \\
\thanks{\textsuperscript{$\dagger$}Corresponding Author.}
}

% \author{Junli Deng,~\IEEEmembership{Student Member,~IEEE}
% \thanks{Junli Deng is with the Department of Computer Science, XYZ University, City, Country (e-mail: junli.deng@xyz.edu).}
% \thanks{Manuscript received November 25, 2024; revised Month Day, Year.}
% }

\maketitle

\begin{abstract}
Reconstructing intricate, ever-changing environments remains a central ambition in computer vision—yet existing solutions often crumble before the complexity of real-world dynamics. We present \textbf{DynaSplat}, an approach that extends Gaussian Splatting to dynamic scenes by integrating dynamic-static separation and hierarchical motion modeling. First, we classify scene elements as static or dynamic through a novel fusion of deformation offset statistics and 2D motion flow consistency, refining our spatial representation to focus precisely where motion matters. We then introduce a hierarchical motion modeling strategy that captures both coarse global transformations and fine-grained local movements, enabling accurate handling of intricate, non-rigid motions. Finally, we integrate physically-based opacity estimation to ensure visually coherent reconstructions, even under challenging occlusions and perspective shifts. Extensive experiments on challenging datasets reveal that DynaSplat not only surpasses state-of-the-art alternatives in accuracy and realism but also provides a more intuitive, compact, and efficient route to dynamic scene reconstruction.
\end{abstract}


\section{Introduction}
\label{sec:intro}

Reconstructing dynamic scenes—a setting where objects move independently, appearances shift, and occlusions frequently arise—is a fundamental task in computer vision and robotics, with applications spanning autonomous driving, immersive telepresence, and augmented reality. To navigate such environments, an intelligent system must form an accurate, temporally coherent representation of the world. Yet, this requirement collides with the complexity of dynamic scenes: traditional methods such as Structure-from-Motion (SfM)~\cite{hartley2004multiple} and SLAM~\cite{durrant2006simultaneous1} are rooted in static-world assumptions and frequently yield artifacts when confronted with freely moving objects. Neural Radiance Fields (NeRF)~\cite{mildenhall2020nerf} and their extensions~\cite{park2021nerfies,yang2023freenerf,LIU2025129653} have broadened our representational toolkit, enabling high-quality synthesis and reconstruction. Still, extending these approaches to dynamic settings often demands per-scene training, faces increased computational loads, and struggles to flexibly handle heterogeneous motion patterns.

Recent advances show promise in addressing these complexities. Gaussian Splatting~\cite{kerbl2023} stands out by representing scenes as sets of 3D Gaussian primitives, enabling more efficient optimization and real-time rendering. However, when moving from static to dynamic scenes, new obstacles emerge. The crux lies in effectively isolating genuinely dynamic elements from stable backgrounds and modeling their motions with both accuracy and efficiency. A plausible hypothesis is that dynamic and static components follow distinct motion distributions. If we can robustly detect and treat dynamic elements independently, we can focus our modeling effort on the regions of change, paving the way to more precise and adaptive representations. Moreover, dynamic scenes often contain multi-scale motion phenomena, from coarse global transformations (e.g., object-level shifts) to subtle local deformations (e.g., shape changes and articulations). Ignoring this hierarchical structure may lead to oversimplifications and reduced fidelity.

To address these challenges, we propose \textbf{DynaSplat}, a method that augments Gaussian Splatting with tailored mechanisms for dynamic scene reconstruction. Our approach first identifies dynamic Gaussians by analyzing the temporal variance of Gaussian deformation offsets (3D geometry signals) and corroborating them with pixel-level optical flow consistency (2D cues). Unlike existing methods~\cite{Motion-aware,zhu2024motiongs,Choy2024Unsupervised3P} that rely on optical flow in isolation, this integration achieves a more reliable distinction of moving Gaussians. Next, we introduce a hierarchical motion modeling strategy, which combines multiple learned motion modes through separate MLPs while regularizing each Gaussian’s motion with its neighbors, enabling the capture of specific motion patterns while maintaining global smoothness. Finally, we incorporate a physically-based opacity computation that makes opacity dependent on the camera-Gaussian angle and distance. This approach better handles surfaces and semi-transparent objects compared to fixed opacity settings in previous work. Additionally, it provides valuable insights for other 3D Gaussian Splatting tasks. Our contributions are threefold:

\begin{itemize}
    \item \textbf{Dynamic-Static Separation}: We develop a robust method to classify Gaussians into dynamic or static sets by integrating per-Gaussian temporal variance and pixel-level optical flow consistency.
    \item \textbf{Coarse-to-Fine Motion Decomposition}: We introduce a structured decomposition of object motions into coarse and fine components, capturing complex dynamics more effectively.
    \item \textbf{Physically-Based Opacity Computation}: We design an opacity calculation guided by physical principles, improving occlusion resolution and depth ordering for more coherent reconstructions.
\end{itemize}

\section{Related Work}
\label{sec:related_work}

Reconstructing dynamic scenes has long challenged traditional methods like multi-view stereo~\cite{seitz1999photorealistic}, SLAM~\cite{durrant2006simultaneous1}, and depth estimation~\cite{eigen2014depth}, as they typically assume static environments. Neural rendering techniques, pioneered by NeRF~\cite{mildenhall2020nerf}, have advanced novel view synthesis and scene representation. While subsequent works extended NeRF to dynamic contexts (e.g., D-NeRF~\cite{pumarola2021dnerf}, NSFF~\cite{li2021nsff}, K-Planes~\cite{fridovich2023k}), these methods remain computationally intensive.

To achieve efficiency, Gaussian Splatting~\cite{kerbl2023} uses 3D Gaussians as primitives for real-time scene rendering. While follow-up~\cite{wu20234dgaussians,guo2024mixed,shaw2024swings,yang2024deformable,xu2024event} efforts have explored dynamic scenarios, there remains a gap in effectively separating static and dynamic elements, and in hierarchically modeling complex object motions.

Our work addresses these limitations by incorporating dynamic-static separation through offset analysis, hierarchical motion modeling, and physically-based opacity computation within the Gaussian Splatting framework. This integrated approach enables more accurate and efficient reconstruction of dynamic scenes.

\section{Proposed Method}
\label{sec:proposed_method}

Our method is built upon 3D Gaussian Splatting (3DGS)~\cite{kerbl2023}. In this section, we first provide a brief overview of 3D Gaussian Splatting, and then detail our three main contributions: dynamic-static separation, physically-based opacity computation, and hierarchical motion modeling.

\subsection{Preliminary: 3D Gaussian Splatting}

3D Gaussian Splatting (3DGS)~\cite{kerbl2023} represents scenes using 3D Gaussian functions as rendering primitives. Each Gaussian is defined by a mean position $\boldsymbol{\mu}$, covariance matrix $\Sigma$, opacity $\sigma$, and view-dependent color $\boldsymbol{c}$, with the function:
\begin{equation}
    G(\mathbf{x}) = e^{ -\frac{1}{2} (\mathbf{x} - \boldsymbol{\mu})^\top \Sigma^{-1} (\mathbf{x} - \boldsymbol{\mu}) }.
\end{equation}
The covariance $\Sigma = R S S^\top R^\top$ is constructed from a rotation matrix $R$ and a diagonal scaling matrix $S$.

During rendering, 3D Gaussians are projected onto the image plane, with mean $\boldsymbol{\mu}$ becoming $\boldsymbol{\mu}_{2D}$ and $\Sigma$ transforming into a 2D covariance $\Sigma'$. Pixel color $C(\mathbf{x})$ is computed as:
\begin{equation}
    C(\mathbf{x}) = \sum_{i \in N(\mathbf{x})} \boldsymbol{c}_i \alpha_i(\mathbf{x}) \prod_{j=1}^{i-1} (1 - \alpha_j(\mathbf{x})),
\end{equation}
where $\alpha_i(\mathbf{x}) = \sigma_i \exp \left( -\frac{1}{2} (\mathbf{x} - \boldsymbol{\mu}_{2D,i})^\top \Sigma'^{-1}_i (\mathbf{x} - \boldsymbol{\mu}_{2D,i}) \right)$.

In practice, $\Sigma$ is parameterized by $R$ and $S$ for optimization, and $\boldsymbol{c}$ uses spherical harmonics for view-dependent effects. A Gaussian is thus represented as $G\{\boldsymbol{\mu}, R, S, \sigma, \boldsymbol{c}\}$.

\subsection{Dynamic-Static Separation}

\begin{figure*}[!t]
\centering
   \includegraphics[width=1.0\textwidth]{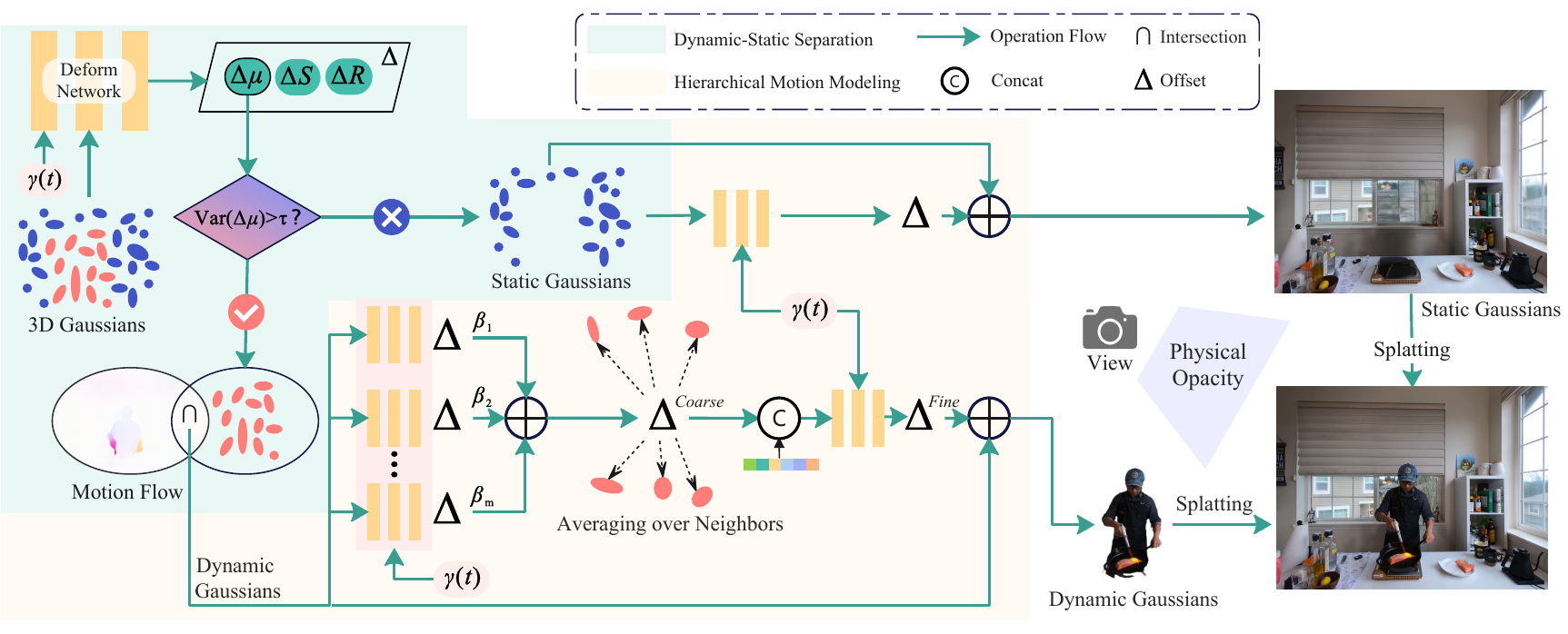}
% \rule{\textwidth}{8cm}  % Placeholder for the pipeline figure
\caption{Overview of DynaSplat. Our method first performs dynamic-static separation through offset variance analysis and 2D motion flow consistency verification. Static Gaussians are processed directly, while dynamic Gaussians undergo hierarchical motion modeling with coarse-to-fine decomposition. The coarse component captures global motion trends through neighbor averaging, while the fine component handles local deformations. Finally, physically-based opacity computation ensures realistic rendering.}
\label{fig_pipeline}
\end{figure*}

Our method employs a canonical and deformation network (DeformNet) architecture. As shown in Fig.~\ref{fig_pipeline}, the DeformNet learns the offsets of position $\Delta \boldsymbol{\mu}_t$, scale $\Delta S_t$, and rotation $\Delta R_t$ for each Gaussian point at each time frame $t$, allowing for flexible modeling of dynamic motions:
%\vspace{-0.1cm}
\begin{align}
\boldsymbol{\mu}_t &= \boldsymbol{\mu}_c + \Delta \boldsymbol{\mu}_t, \\
S_t &= S_c + \Delta S_t, \\
R_t &= R_c + \Delta R_t,
\end{align}
where the subscript $c$ denotes the canonical values.

During the early training stages, we train the basic canonical and DeformNet architecture without differentiating between static and dynamic Gaussians. This allows the model to capture the general structure and motions in the scene.

After the initial training phase, we proceed to identify dynamic Gaussians using the following methods.

\paragraph{Offset Variance Analysis}

We compute the variance of the position offsets $\Delta \boldsymbol{\mu}_t$ for each Gaussian over all frames $t = 1, \dots, T$. The variance of the position offsets is calculated as:
%%\vspace{-0.1cm}
\begin{equation}
\text{Var}(\Delta \boldsymbol{\mu}) = \frac{1}{T} \sum_{t=1}^{T} \left\| \Delta \boldsymbol{\mu}_t - \overline{\Delta \boldsymbol{\mu}} \right\|^2,
\end{equation}
where $\overline{\Delta \boldsymbol{\mu}}$ is the mean of the position offsets over all frames:
%\vspace{-0.1cm}
\begin{equation}
\overline{\Delta \boldsymbol{\mu}} = \frac{1}{T} \sum_{t=1}^{T} \Delta \boldsymbol{\mu}_t.
\end{equation}

Gaussians whose position offset variance exceeds a predefined threshold $\tau$ are considered as candidates for being dynamic:

\begin{equation}
\mathcal{G}_{\text{variance}} = \left\{ g \in \mathcal{G} \ \bigg| \ \text{Var}(\Delta \boldsymbol{\mu}_g) > \tau \right\}.
\end{equation}

\paragraph{2D Motion Flow Consistency}

To verify whether these candidate Gaussians are truly dynamic, we check the consistency of their projected positions with a 2D motion flow map~\cite{dong2024memflow}. Specifically, for each candidate dynamic Gaussian $g \in \mathcal{G}_{\text{variance}}$, we project its updated positions $\boldsymbol{\mu}_{t}$ onto the image plane to obtain 2D positions $\mathbf{u}_{g,t}$ at each frame $t$. We then assess whether $\mathbf{u}_{g,t}$ falls within regions of significant motion in the 2D motion flow map $F^M_{t \to t+1}$:

\begin{equation}
\mathcal{F}_g = \left\{ t \ \bigg| \ \| F^M_{t \to t+1}(\mathbf{u}_{g,t}) \|_2 > \epsilon \right\}.
\end{equation}

If the proportion of frames where the motion flow magnitude exceeds the threshold $\epsilon$ is significant (e.g., more than 50\%), we confirm that the Gaussian is dynamic:

\begin{equation}
\frac{|\mathcal{F}_g|}{T} > \gamma,
\end{equation}
where $T$ is the total number of frames and $\gamma$ is a predefined ratio (e.g., $\gamma = 0.5$). The final set of dynamic Gaussians is then defined as:

\begin{equation}
\mathcal{G}_{\text{dynamic}} = \left\{ g \in \mathcal{G}_{\text{variance}} \ \bigg| \ \frac{|\mathcal{F}_g|}{T} > \gamma \right\}.
\end{equation}

The remaining Gaussians form the static set $\mathcal{G}_{\text{static}} = \mathcal{G} \setminus \mathcal{G}_{\text{dynamic}}$.

\paragraph{Obtaining the 2D Motion Flow Map}

Following MotionGS~\cite{zhu2024motiongs}, we first estimate optical flow using an off-the-shelf method, then remove the camera-induced component derived from known camera poses and rendered depth maps. The resulting motion flow map reveals regions of actual object movement.

\paragraph{Final Dynamic Gaussian Set}

The final set of dynamic Gaussians is defined as:

\begin{equation}
\mathcal{G}_{\text{dynamic}} = \left\{ g \in \mathcal{G}_{\text{variance}} \ \bigg| \ \frac{|\mathcal{F}_g|}{T} > \gamma \right\}.
\end{equation}

Gaussians not included in $\mathcal{G}_{\text{dynamic}}$ are processed with a lightweight three MLP, retaining parameters from the early training stages. Gaussians in $\mathcal{G}_{\text{dynamic}}$ undergo hierarchical motion modeling using Adaptive Motion Networks, which utilize multiple motion modes and blending parameters.

\subsection{Hierarchical Motion Modeling}

After the early training stages, we employ a hierarchical motion modeling approach to capture detailed motions of dynamic objects. In training, for static points, we use a lightweight three-layer MLP, keeping the parameters from the early training stages. For dynamic points, we utilize multiple \textbf{Adaptive Motion Networks}, which are also initialized with the parameters from the early training stages. 

As shown in Fig.~\ref{fig_pipeline}, our hierarchical motion modeling captures both global motion trends and local motion details:

\subsubsection{Adaptive Motion Networks}

Adaptive Motion Networks model dynamic motions by blending multiple MLP outputs with learnable parameters. We encode position and time, and let the blending parameters $\{\beta_m\}$ determine how much each motion mode contributes. The final offsets are given by a single weighted sum:

{\small
\begin{equation}
    \Delta \boldsymbol{\mu}(t), \;\Delta R(t), \;\Delta S(t) \;=\; 
    \sum_{m=1}^M \beta_m \cdot \text{MLP}_m\bigl(\gamma(\boldsymbol{\mu}), \;\gamma(t), \;S_c, \;R_c\bigr),
\end{equation}}
where $\gamma(\cdot)$ denotes positional encoding, $S_c$ and $R_c$ are additional input features, and $\{\beta_m\}$ are the learnable blending parameters. This formulation enables the network to capture complex dynamic behaviors through a flexible combination of different motion modes.

\subsubsection{Coarse-to-Fine Motion Decomposition}

To handle position, orientation, and scale simultaneously, we group these parameters into a single vector. For each Gaussian $g$, let:
\[
    \Delta \mathbf{X}_g = (\Delta \boldsymbol{\mu}_g, \;\Delta R_g, \;\Delta S_g).
\]

We first extract a coarse estimate by averaging over neighboring Gaussians:
\[
    \Delta \mathbf{X}_g^{\text{coarse}} = \frac{1}{|\mathcal{N}_g|} \sum_{i \in \mathcal{N}_g} \Delta \mathbf{X}_i.
\]

Next, we apply a multi-layer perceptron (MLP) to refine this coarse estimate using a learnable feature vector $\boldsymbol{f}_g$ (included only for dynamic points):
\[
    \Delta \mathbf{X}_g^{\text{fine}} = \text{MLP}(\Delta \mathbf{X}_g^{\text{coarse}}, \;\boldsymbol{f}_g).
\]

The final motion parameters are then:
\[
    \Delta \mathbf{X}_g = \Delta \mathbf{X}_g^{\text{coarse}} \;+\; \Delta \mathbf{X}_g^{\text{fine}}.
\]

This hierarchical approach captures both global trends and fine-grained details, improving the fidelity of dynamic scene reconstruction.

% \subsubsection{Advantages of Hierarchical Motion Modeling}

% Our hierarchical motion modeling offers several benefits:

% \begin{itemize}
%     \item \textbf{Enhanced Generalization}: Capturing global motion trends reduces overfitting to local variations.
%     \item \textbf{Improved Detail Modeling}: The adaptive network learns complex local motions, increasing reconstruction accuracy.
%     \item \textbf{Computational Efficiency}: Hierarchical processing reduces computational load compared to modeling all Gaussians individually.
% \end{itemize}

\begin{figure}[t]
    \centering
    % \rule{0.9\linewidth}{5cm}  % Placeholder for the opacity computation figure
    \includegraphics[width=0.6\linewidth]{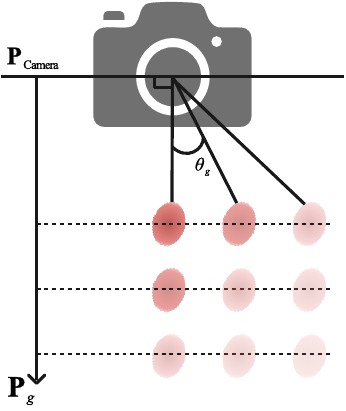}
    \caption{Illustration of physically-based opacity computation.}
    \label{fig:opacity_computation}
\end{figure}
%\vspace{-0.1cm}

\subsection{Physically-Based Opacity Computation}

Accurate opacity handling is crucial for realistic rendering. We introduce a physically-based method for computing the opacity of Gaussians, considering their distance and viewing angle, which is shown is Fig.~\ref{fig:opacity_computation}.

\subsubsection{Physical Modeling of Opacity}

We define the opacity $\alpha_g$ of a Gaussian $g$ as inversely proportional to the square of its distance from the camera and modulated by the viewing angle $\theta_g$ between the Gaussian's normal $\mathbf{n}_g$ and the viewing direction $\mathbf{v}_g$:

\begin{equation}
    \alpha_g = \alpha_0 \cdot \frac{\cos \theta_g}{\|\mathbf{P}_g - \mathbf{P}_{\text{camera}}\|_2^2},
\end{equation}
where $\alpha_0$ is an initial opacity constant, $\mathbf{P}_g$ is the Gaussian center, and $\cos \theta_g = \mathbf{n}_g^\top \mathbf{v}_g$.

\subsubsection{Temporal Importance Filtering}

Unlike the original 3DGS which regularly reduces opacity and clips low-transparency Gaussians, we eliminate the reset opacity operation to avoid excessive coupling between canonical and deformation spaces. To address the resulting floaters, we propose \textbf{Temporal Importance Filtering}.

This involves calculating the importance of each Gaussian point to each training viewpoint at every timestamp. Gaussians with importance below a certain threshold are clipped, effectively reducing floater issues. For a Gaussian point $g_i$, the importance $w_i$ is calculated as follows:

\begin{equation}
    w_i = \max_{I\in\mathcal{I}, \mathbf{x}\in I, t\in T} \left( \alpha_i(\mathbf{x}|t) \prod_{j=1}^{i-1}(1 - \alpha_j(\mathbf{x}|t)) \right),
\end{equation}
where $\mathcal{I}$ represents the images from all training views, $\alpha_i(\mathbf{x}|t)$ is the opacity of $g_i$ at time $t$ for pixel $\mathbf{x}$, and $T$ represents the set of query times. We prune Gaussians when their importance satisfies $w_i < \tau$, where $\tau$ is a predefined threshold.

In previous work~\cite{guo2024motion,lin2024gaussian,yang2024deformable,icml2024-sp-gs}, Gaussian points were pruned based on spatial attributes like transparency and volume. In contrast, we prune points based on their importance across all training views in the temporal domain. This method effectively eliminates artifacts that are suspended in the air and were not captured by the training views.

\subsection{Loss Function}

We use a combination of loss functions to supervise the training process.:

\begin{itemize}
    \item \textbf{Reconstruction Loss}: Combines mean squared error (MSE) and structural similarity index (SSIM) between rendered images $\hat{I}_t$ and ground truth images $I_t$:
    \begin{equation}
        \mathcal{L}_{\text{recon}} = (1 - \lambda) \| \hat{I}_t - I_t \|_1 + \lambda \mathcal{L}_{\mathrm{SSIM}}(\hat{I}_t, I_t),
    \end{equation}
    where $\lambda$ balances the two terms.
    \item \textbf{Total Variation Loss}~\cite{wu20234dgaussians}: Encourages spatial smoothness by penalizing abrupt parameter changes between neighboring Gaussians.
\end{itemize}

% \subsubsection{Data Preprocessing}

% To improve the correspondence between pixels and Gaussians, we perform:

% \begin{itemize}
%     \item \textbf{Estimation of Ground Truth Optical Flow}: Compute optical flow between consecutive frames to assist in dynamic-static separation and motion modeling.
%     \item \textbf{Depth Estimation}: Use depth estimation networks to obtain depth maps when unavailable.
%     \item \textbf{Normalization}: Normalize input images and depth maps to maintain numerical stability.
% \end{itemize}

\section{Experiments}
\label{sec:experiments}

We evaluate our method on two dynamic scene datasets: the N3DV dataset~\cite{li2022neural} containing complex motions and challenging lighting conditions, and the D-NeRF dataset~\cite{pumarola2021dnerf} providing synthetic scenes with ground truth geometry and motion fields. For quantitative evaluation, we use Peak Signal-to-Noise Ratio (PSNR), Structural Similarity Index (SSIM), and Learned Perceptual Image Patch Similarity (LPIPS)~\cite{zhang2018unreasonable} to assess image quality, structural similarity, and perceptual similarity, respectively.

\subsection{Implementation Details}

Our model is implemented in PyTorch and trained on an NVIDIA RTX 3090 GPU for 150k iterations. The learning rate for static Gaussians follows the 3DGS settings, while the deformation networks' learning rate decays from $8 \times 10^{-4}$ to $1.6 \times 10^{-6}$. During the first 3k iterations (early training stage), only one DeformNet is optimized. Dynamic-static separation is introduced after 6k iterations with thresholds $\tau = 0.01$, $\epsilon = 1.0$, and $\gamma = 0.5$. Our Adaptive Motion Networks utilize $M = 4$ motion modes. Temporal Importance Filtering is activated after 20k iterations with a pruning threshold of $\tau = 0.02$. For the reconstruction loss, we set the initial opacity constant to $\alpha_0 = 1.0$ and use a balance parameter of $\lambda = 0.1$. The entire training process takes approximately 2 hours on a single GPU.

\subsection{Results}

Our method achieves superior reconstruction quality compared to existing state-of-the-art (SOTA) methods. Table~\ref{tab_results_synthetic} and Table~\ref{tab_results_real} present quantitative comparisons on the D-NeRF and N3DV Video datasets, respectively.

\begin{table}[!tbp]
\caption{Quantitative Comparison on D-NeRF Dataset}
\centering
\renewcommand{\arraystretch}{0.9}  % Reduce row spacing
\footnotesize
\begin{tabular}{@{}lccc@{}}
\toprule
Method & PSNR $\uparrow$ & SSIM $\uparrow$ & LPIPS $\downarrow$ \\
\midrule
DyNeRF~\cite{li2022neural} & 29.58 & - & 0.080 \\
StreamRF~\cite{li2022streaming} & 28.16 & 0.850 & 0.310 \\
HyperReel~\cite{attal2023hyperreel} & 30.36 & 0.920 & 0.170 \\
NeRFPlayer~\cite{song2023nerfplayer} & 30.69 & - & 0.110 \\
K-Planes~\cite{fridovich2023k} & 31.05 & 0.950 & 0.040 \\
4D-GS~\cite{wu20234dgaussians} & 31.80 & 0.958 & 0.032 \\
Def-3D-Gauss~\cite{yang2024deformable} & 32.00 & 0.960 & 0.030 \\
4D-Rotor-Gauss~\cite{duan:2024:4drotorgs} & 34.25 & 0.962 & 0.048 \\
\textbf{Ours} & \textbf{34.39} & \textbf{0.965} & \textbf{0.034} \\
\bottomrule
\end{tabular}
\label{tab_results_synthetic}
\end{table}

\begin{table}[!htbp]
\caption{Quantitative Comparison on the N3DV Dataset}
\centering
\renewcommand{\arraystretch}{0.9}  % Reduce row spacing
\footnotesize
\begin{tabular}{@{}lccc@{}}
\toprule
Method & PSNR $\uparrow$ & SSIM $\uparrow$ & LPIPS $\downarrow$ \\
\midrule
DyNeRF~\cite{li2022neural} & 29.58 & - & 0.080 \\
StreamRF~\cite{li2022streaming} & 28.16 & 0.850 & 0.310 \\
HyperReel~\cite{attal2023hyperreel} & 30.36 & 0.920 & 0.170 \\
NeRFPlayer~\cite{song2023nerfplayer} & 30.69 & - & 0.110 \\
K-Planes~\cite{fridovich2023k} & 30.73 & 0.930 & 0.070 \\
MixVoxels~\cite{Wang2023ICCV} & 30.85 & 0.960 & 0.210 \\
4D-GS~\cite{yangreal} & 29.91 & 0.928 & 0.168 \\
4D-Rotor-Gauss~\cite{duan:2024:4drotorgs} & \textbf{31.80} & 0.935 & 0.142 \\
\textbf{Ours} & 31.68 & \textbf{0.967} & \textbf{0.069} \\
\bottomrule
\end{tabular}
\label{tab_results_real}
\end{table}

\begin{figure}[!t]
\centering
\includegraphics[width=3.4in]{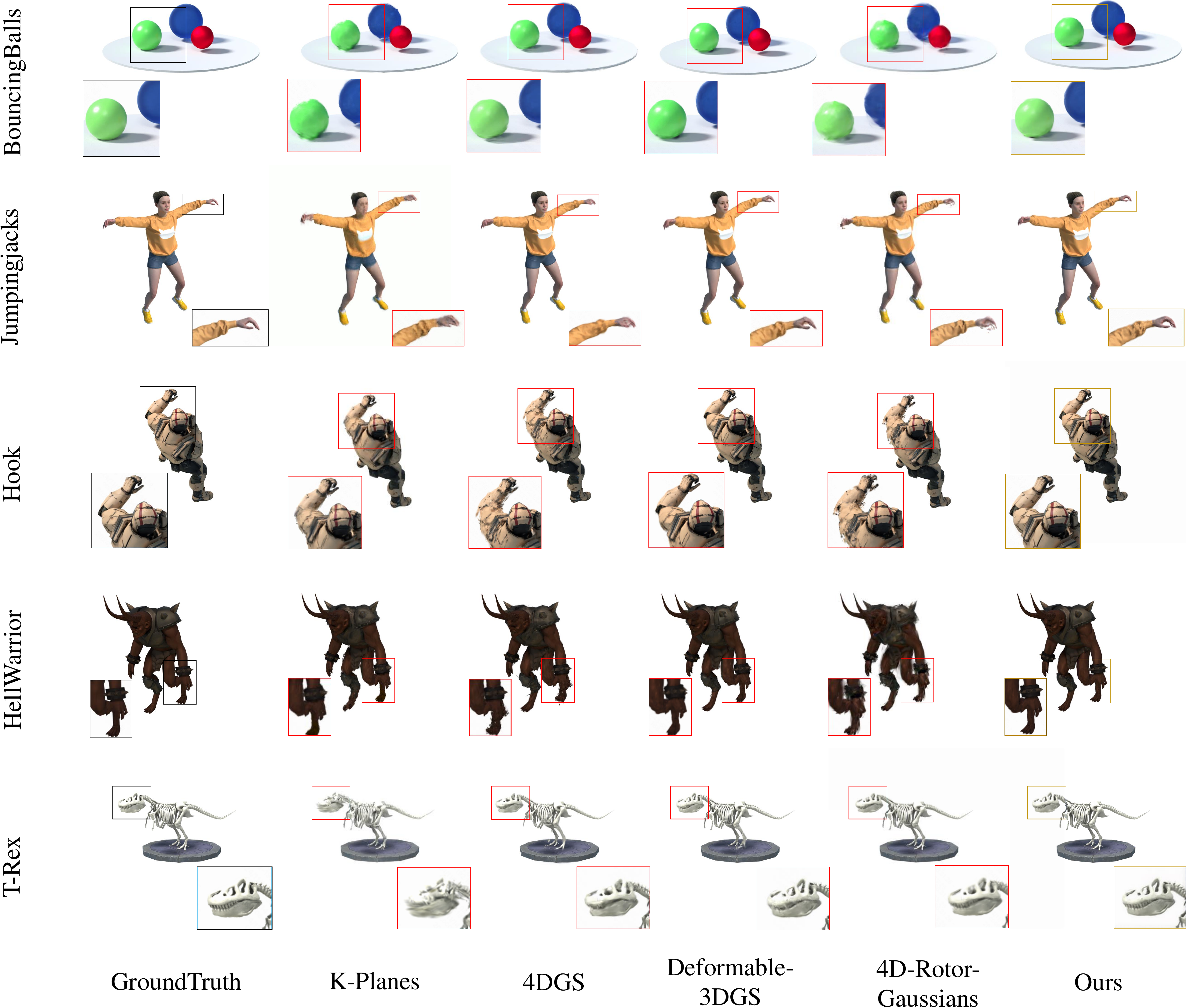}
% \rule{\linewidth}{4cm}  % Placeholder for qualitative results
\caption{Comparison of scenes from D-NeRF dataset.}
\label{fig_results_dnerf}
\end{figure}

\begin{figure}[!t]
\centering
\includegraphics[width=3.5in]{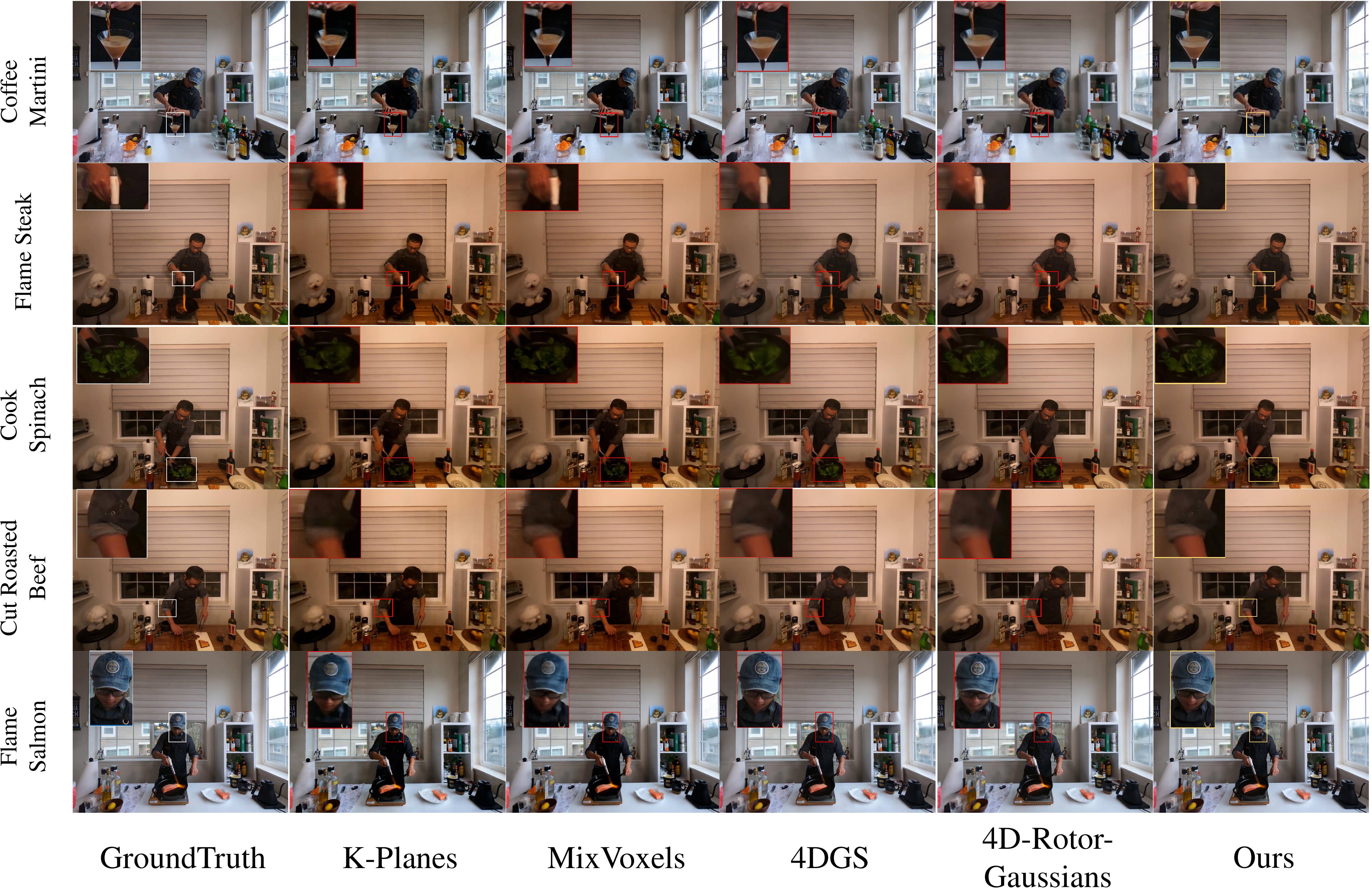}
% \rule{\linewidth}{4cm}  % Placeholder for qualitative results
\caption{Comparison of scenes from N3DV dataset.}
\label{fig_results_dynerf}
\end{figure}

As shown in Fig.~\ref{fig_results_dnerf} and Fig.~\ref{fig_results_dynerf}, our method achieves superior reconstruction quality with sharper details and better accuracy on dynamic objects.

\subsection{Application to High-Fidelity Head Avatar Reconstruction}
To rigorously evaluate the versatility and robustness of our framework beyond general dynamic scenes, we applied DynaSplat to the highly demanding task of dynamic head avatar reconstruction. This task is notoriously challenging as it requires not only capturing the global motion of the head but also modeling the subtle, non-rigid deformations of facial features with exceptional fidelity to preserve personal identity and convey emotion accurately. Our method proves to be remarkably effective in this domain by leveraging its core architectural principles.

The success is primarily attributed to our \textbf{Hierarchical Motion Modeling}. Human head dynamics are a perfect example of composite motion, which our coarse-to-fine decomposition is ideally suited to handle.
\begin{itemize}
    \item The \textbf{coarse motion component} effectively captures the large-scale, rigid transformations of the head, such as nodding, shaking, and tilting. By averaging motion information from neighboring Gaussians, it establishes a stable and coherent baseline for the overall head pose in each frame.
    \item The \textbf{fine motion component}, powered by our Adaptive Motion Networks, then focuses exclusively on the residual, non-rigid deformations. This allows the model to learn the intricate, localized movements of facial muscles—the stretch of the lips during speech, the crinkling of skin around the eyes in a smile, and the subtle lift of an eyebrow. This explicit decomposition prevents the model from becoming unstable and allows for a more precise and detailed reconstruction of facial expressions, which is a common failure point for methods with a single motion model.
\end{itemize}

Furthermore, our other key contributions play a crucial role. Although the head is the primary dynamic element, \textbf{Dynamic-Static Separation} efficiently isolates the moving head from any static background elements, focusing computational resources where they are most needed. More critically, the \textbf{Physically-Based Opacity Computation} is instrumental in achieving photorealism. By modulating opacity based on viewing angle and distance, our model produces realistic specular highlights on the skin (e.g., on the forehead and nose) and renders soft, semi-transparent details like hair strands more naturally, avoiding the hard, artificial edges often seen in other rendering techniques.

\begin{figure}[!t]
\centering
\includegraphics[width=3.4in]{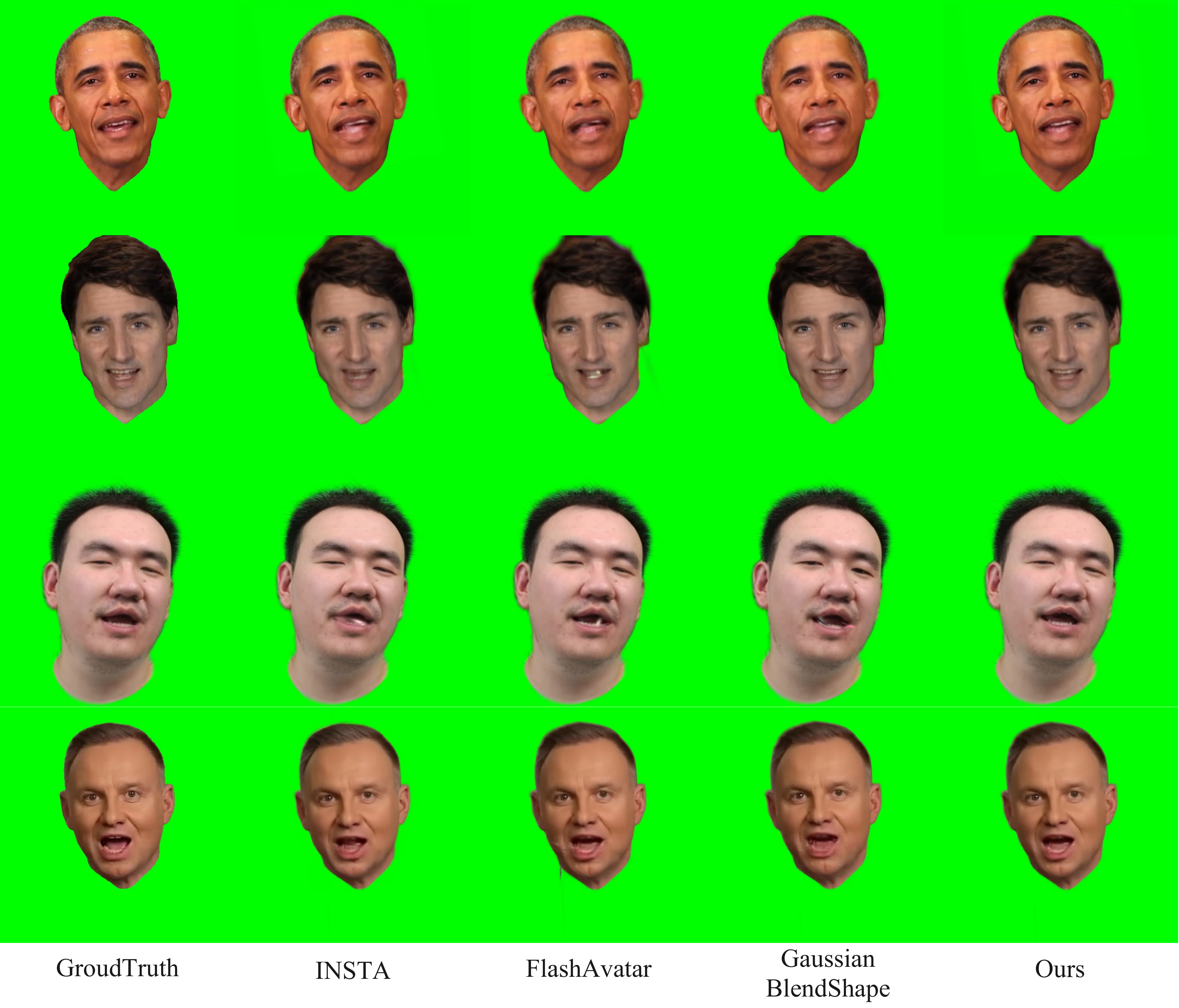}
\caption{Qualitative comparison on dynamic head avatar reconstruction. From left to right: Ground Truth, INSTA, FlashAvatar, Gaussian BlendShape, and Our method (DynaSplat). Our approach demonstrates high-fidelity results comparable to specialized state-of-the-art avatar creation techniques.}
\label{fig:head_avatar_comparison}
\end{figure}

As shown in Fig.~\ref{fig:head_avatar_comparison}, a qualitative comparison with state-of-the-art specialized avatar methods reveals that DynaSplat produces reconstructions that are highly competitive, capturing nuanced expressions with exceptional clarity and realism. This visual evidence is further substantiated by the quantitative results presented in Table~\ref{tab:head_avatar_comparison}. Our method achieves superior performance across all major image quality metrics, including PSNR and LPIPS, while maintaining highly competitive training and inference speeds. This demonstrates that DynaSplat is not just a powerful tool for general dynamic scenes but also a robust and effective solution for specialized, high-fidelity applications like digital avatar creation.

\begin{table}[!t]
\centering
\caption{Quantitative comparisons with state-of-the-art dynamic head avatar reconstruction methods on a benchmark dataset. Our method demonstrates superior performance in reconstruction quality while maintaining competitive efficiency.}
\setlength{\tabcolsep}{4pt} % Adjust column separation
\renewcommand{\arraystretch}{1.1} % Adjust row spacing
\begin{tabular}{@{}lcccccc@{}}
\toprule
Method & MSE $\downarrow$ & SSIM $\uparrow$ & PSNR $\uparrow$ & LPIPS $\downarrow$ & \makecell{Training \\ (Minutes)} & \makecell{Inference \\ (FPS)} \\
\midrule
INSTA~\cite{INSTA:CVPR2023} & 0.0078 & 0.942 & 30.34 & 0.0358 & 12 & 20 \\
FlashAvatar~\cite{xiang2024flashavatar} & 0.0077 & 0.938 & 31.07 & 0.0299 & \textbf{8.2} & 103 \\
GaussianBS~\cite{ma2024gaussianblendshapes} & 0.0080 & 0.935 & 32.52 & 0.0351 & 25 & \textbf{355} \\
\textbf{Ours} & \textbf{0.0071} & \textbf{0.946} & \textbf{33.24} & \textbf{0.0281} & 11 & 324 \\
\bottomrule
\end{tabular}
\label{tab:head_avatar_comparison}
\end{table}

\subsection{Ablation Studies}

We conduct ablation studies on both the D-NeRF and N3DV Video datasets to evaluate each component's contribution, with results shown in Table~\ref{tab_ablation_combined}. The consistent performance degradation observed when removing individual components demonstrates their essential role in our approach.

\begin{table*}[!t]
\caption{Ablation Study Results on D-NeRF and N3DV Video Datasets}
\centering
\renewcommand{\arraystretch}{0.9}  % Reduce row spacing
\footnotesize
\begin{tabular}{lcccccc}
\hline
\multirow{2}{*}{Method} & \multicolumn{3}{c}{D-NeRF Dataset} & \multicolumn{3}{c}{N3DV Video Dataset} \\
\cline{2-7}
& PSNR $\uparrow$ & SSIM $\uparrow$ & LPIPS $\downarrow$ & PSNR $\uparrow$ & SSIM $\uparrow$ & LPIPS $\downarrow$ \\
\hline
Full Model & \textbf{34.39} & \textbf{0.965} & \textbf{0.034} & \textbf{31.68} & \textbf{0.967} & \textbf{0.069} \\
\hline
Without Dynamic-Static Separation & 33.12 & 0.960 & 0.040 & 30.50 & 0.925 & 0.160 \\
Without Hierarchical Motion Modeling & 33.34 & 0.962 & 0.038 & 30.80 & 0.928 & 0.155 \\
Without Physically-Based Opacity Computation & 33.65 & 0.965 & 0.035 & 31.00 & 0.930 & 0.150 \\
\hline
\end{tabular}
\label{tab_ablation_combined}
\end{table*}
% %\vspace{-0.1cm}

\subsubsection{Effect of Dynamic-Static Separation}

\begin{figure}[!t]
\centering
\includegraphics[width=2.5in]{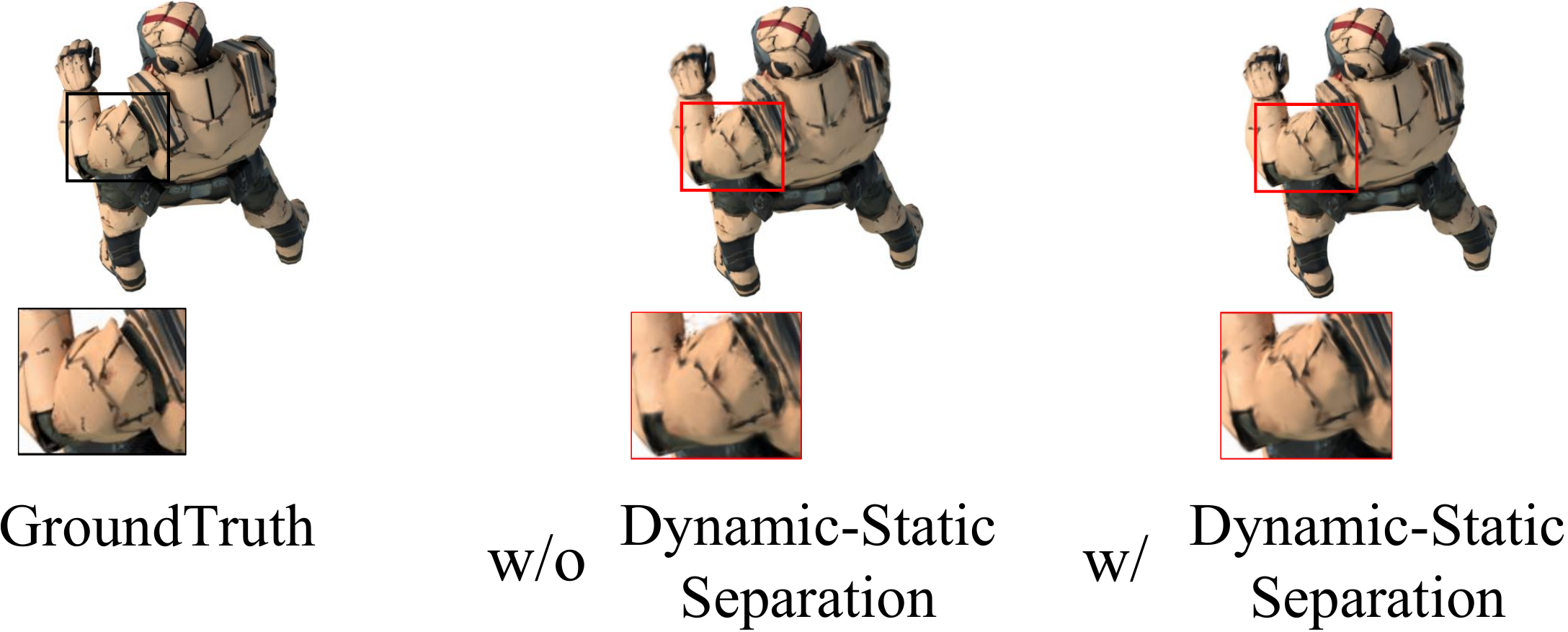}
\caption{Effect of Dynamic-Static Separation.}
\label{fig_ablation1}
\end{figure}

As shown in Fig.~\ref{fig_ablation1}, omitting dynamic-static separation degrades reconstruction quality, especially where dynamic and static elements interact, highlighting its importance in scene modeling.

\subsubsection{Effect of Hierarchical Motion Modeling}

\begin{figure}[!t]
\centering
\includegraphics[width=2.4in]{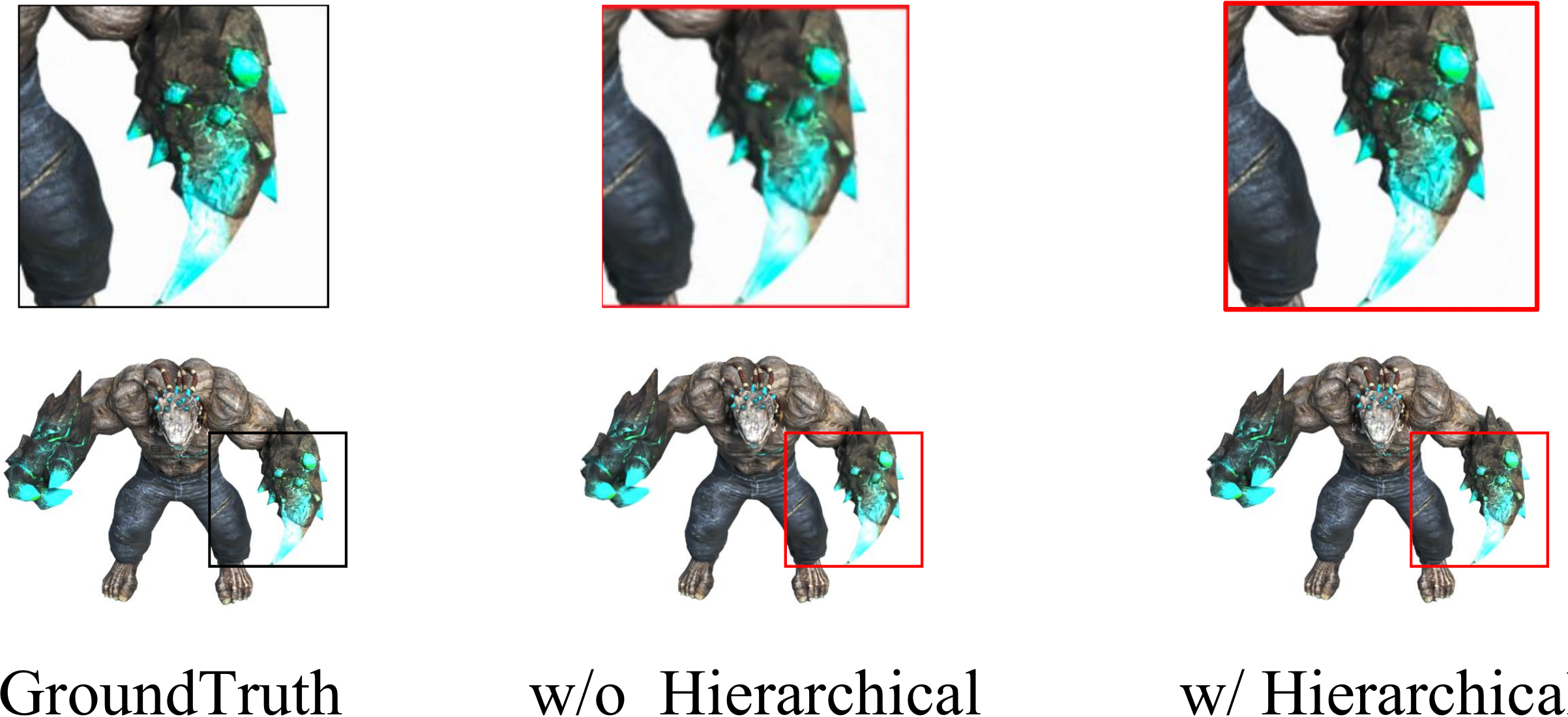}
\caption{Effect of Hierarchical Motion Modeling.}
\label{fig_ablation2}
\end{figure}

Fig.~\ref{fig_ablation2} illustrates that replacing hierarchical motion modeling with a single-level model results in less detailed reconstructions. Our hierarchical approach better preserves sharper textures and captures both global motion patterns and local details.

\subsubsection{Effect of Physically-Based Opacity Computation}

\begin{figure}[!t]
\centering
\includegraphics[width=2.6in]{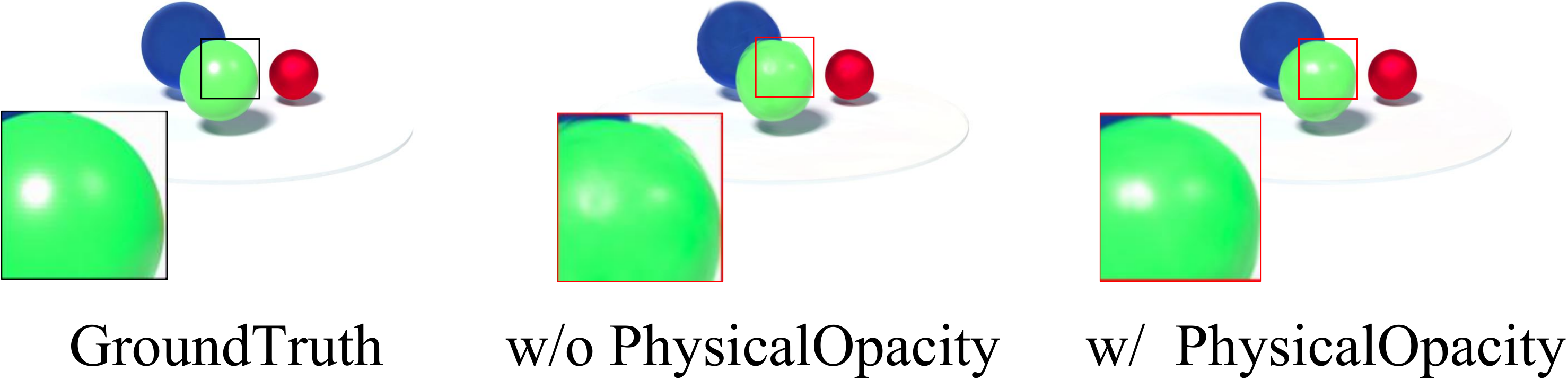}
\caption{Effect of Physically-Based Opacity Computation.}
\label{fig_ablation3}
\end{figure}

As demonstrated in Fig.~\ref{fig_ablation3}, our physically-based opacity computation significantly improves the rendering of specular highlights and reflections, especially on spherical objects, leading to more realistic results that better match the ground truth.

\section{Conclusion}
\label{sec:conclusion}

We have presented DynaSplat, a novel dynamic scene reconstruction method that effectively addresses the challenges of modeling complex, real-world dynamic environments through three key innovations: dynamic-static separation using offset variance analysis and motion flow consistency, hierarchical motion modeling for capturing both global and local motion patterns, and physically-based opacity computation for enhanced rendering realism. Our extensive experiments demonstrate that DynaSplat outperforms existing state-of-the-art methods, with ablation studies revealing significant contributions from each component. While our current results are promising, future work could explore handling multiple interacting dynamic objects, real-time optimization, and physics-based constraints.

% \section*{Acknowledgment}

% The author would like to thank the XYZ Lab for valuable discussions and the anonymous reviewers for their constructive comments.

\bibliographystyle{IEEEtran}
\bibliography{references}

\begin{thebibliography}{10}
\providecommand{\url}[1]{#1}
\csname url@samestyle\endcsname
\providecommand{\newblock}{\relax}
\providecommand{\bibinfo}[2]{#2}
\providecommand{\BIBentrySTDinterwordspacing}{\spaceskip=0pt\relax}
\providecommand{\BIBentryALTinterwordstretchfactor}{4}
\providecommand{\BIBentryALTinterwordspacing}{\spaceskip=\fontdimen2\font plus
\BIBentryALTinterwordstretchfactor\fontdimen3\font minus \fontdimen4\font\relax}
\providecommand{\BIBforeignlanguage}[2]{{%
\expandafter\ifx\csname l@#1\endcsname\relax
\typeout{** WARNING: IEEEtran.bst: No hyphenation pattern has been}%
\typeout{** loaded for the language `#1'. Using the pattern for}%
\typeout{** the default language instead.}%
\else
\language=\csname l@#1\endcsname
\fi
#2}}
\providecommand{\BIBdecl}{\relax}
\BIBdecl

\bibitem{hartley2004multiple}
R.~Hartley and A.~Z.~E. {Z}, \emph{Multiple View Geometry in Computer Vision}, 2nd~ed.\hskip 1em plus 0.5em minus 0.4em\relax Cambridge University Press, 2004.

\bibitem{durrant2006simultaneous1}
H.~Durrant-Whyte and T.~Bailey, ``Simultaneous localization and mapping (slam): Part i the essential algorithms,'' \emph{IEEE Robotics \& Automation Magazine}, vol.~13, no.~2, pp. 99--110, 2006.

\bibitem{mildenhall2020nerf}
B.~Mildenhall, P.~P. Srinivasan, M.~Tancik, J.~T. Barron, R.~Ramamoorthi, and R.~Ng, ``Nerf: Representing scenes as neural radiance fields for view synthesis,'' in \emph{ECCV}, 2020, pp. 405--421.

\bibitem{park2021nerfies}
J.~J. Park and et~al., ``Nerfies: Deformable neural radiance fields,'' in \emph{Proceedings of the IEEE/CVF International Conference on Computer Vision (ICCV)}, 2021, pp. 4480--4490.

\bibitem{yang2023freenerf}
J.~Yang, M.~Pavone, and Y.~Wang, ``Freenerf: Improving few-shot neural rendering with free frequency regularization,'' in \emph{CVPR}, 2023.

\bibitem{LIU2025129653}
\BIBentryALTinterwordspacing
J.~Liu, H.~Cheng, S.~Wang, F.~Zhao, and M.~Li, ``Nerf dynamic scene reconstruction based on motion, semantic information and inpainting,'' \emph{Neurocomputing}, vol. 630, p. 129653, 2025. [Online]. Available: \url{https://www.sciencedirect.com/science/article/pii/S092523122500325X}
\BIBentrySTDinterwordspacing

\bibitem{kerbl2023}
B.~Kerbl, A.~Ruiz, T.~Leimk{\"u}hler, C.~Buehler, S.~Saito, A.~Vedaldi, T.~Weyrich, H.-P. Seidel, B.~Bickel, M.~Rittig \emph{et~al.}, ``3d gaussian splatting for real-time radiance field rendering,'' \emph{ACM Trans. Graph. (SIGGRAPH)}, vol.~42, no.~4, 2023.

\bibitem{Motion-aware}
Z.~Guo, W.~Zhou, L.~Li, M.~Wang, and H.~Li, ``Motion-aware 3d gaussian splatting for efficient dynamic scene reconstruction,'' \emph{IEEE Transactions on Circuits and Systems for Video Technology}, pp. 1--1, 2024.

\bibitem{zhu2024motiongs}
R.~Zhu, Y.~Liang, H.~Chang, J.~Deng, J.~Lu, W.~Yang, T.~Zhang, and Y.~Zhang, ``Motiongs: Exploring explicit motion guidance for deformable 3d gaussian splatting,'' \emph{NeurIPS}, 2024.

\bibitem{Choy2024Unsupervised3P}
\BIBentryALTinterwordspacing
J.~Choy, G.~Cha, H.~Kee, and S.~Oh, ``Unsupervised 3d part decomposition via leveraged gaussian splatting,'' \emph{2024 IEEE/RSJ International Conference on Intelligent Robots and Systems (IROS)}, pp. 2647--2652, 2024. [Online]. Available: \url{https://api.semanticscholar.org/CorpusID:275021195}
\BIBentrySTDinterwordspacing

\bibitem{seitz1999photorealistic}
S.~M. Seitz and C.~R. Dyer, ``Photorealistic scene reconstruction by voxel coloring,'' \emph{International journal of computer vision}, vol.~35, pp. 151--173, 1999.

\bibitem{eigen2014depth}
D.~Eigen \emph{et~al.}, ``Depth map prediction from a single image using a multi-scale deep network,'' in \emph{Proceedings of the Advances in Neural Information Processing Systems (NeurIPS)}, 2014, pp. 2366--2374.

\bibitem{pumarola2021dnerf}
A.~Pumarola, D.~Azinovic, R.~Martin-Brualla, G.~Pons-Moll, and M.~Zollh{\"o}fer, ``D-nerf: Neural radiance fields for dynamic scenes,'' in \emph{CVPR}, 2021, pp. 10\,318--10\,327.

\bibitem{li2021nsff}
Z.~Li, S.~Niklaus, N.~Snavely, and O.~Wang, ``Neural scene flow fields for space-time view synthesis of dynamic scenes,'' in \emph{CVPR}, 2021, pp. 6498--6508.

\bibitem{fridovich2023k}
S.~Fridovich-Keil, G.~Meanti, F.~R. Warburg, B.~Recht, and A.~Kanazawa, ``K-planes: Explicit radiance fields in space, time, and appearance,'' in \emph{Proceedings of the IEEE/CVF Conference on Computer Vision and Pattern Recognition}, 2023, pp. 12\,479--12\,488.

\bibitem{wu20234dgaussians}
G.~Wu, T.~Yi, J.~Fang, L.~Xie, X.~Zhang, W.~Wei, W.~Liu, Q.~Tian, and W.~Xinggang, ``4d gaussian splatting for real-time dynamic scene rendering,'' in \emph{CVPR}, 2024.

\bibitem{guo2024mixed}
J.~Guo, T.~Wang, and C.~Wang, ``Mixed 3d gaussian for dynamic scenes representation and rendering,'' in \emph{2024 IEEE International Conference on Multimedia and Expo (ICME)}.\hskip 1em plus 0.5em minus 0.4em\relax IEEE, 2024, pp. 1--6.

\bibitem{shaw2024swings}
R.~Shaw, M.~Nazarczuk, J.~Song, A.~Moreau, S.~Catley-Chandar, H.~Dhamo, and E.~P{\'e}rez-Pellitero, ``Swings: sliding windows for dynamic 3d gaussian splatting,'' in \emph{ECCV}, 2024.

\bibitem{yang2024deformable}
Z.~Yang, X.~Gao, W.~Zhou, S.~Jiao, Y.~Zhang, and X.~Jin, ``Deformable 3d gaussians for high-fidelity monocular dynamic scene reconstruction,'' in \emph{Proceedings of the IEEE/CVF Conference on Computer Vision and Pattern Recognition}, 2024, pp. 20\,331--20\,341.

\bibitem{xu2024event}
W.~Xu, W.~Weng, Y.~Zhang, R.~Xu, and Z.~Xiong, ``Event-boosted deformable 3d gaussians for fast dynamic scene reconstruction,'' \emph{arXiv preprint arXiv:2411.16180}, 2024.

\bibitem{dong2024memflow}
Q.~Dong and Y.~Fu, ``Memflow: Optical flow estimation and prediction with memory,'' in \emph{Proceedings of the IEEE/CVF Conference on Computer Vision and Pattern Recognition}, 2024.

\bibitem{guo2024motion}
Z.~Guo, W.~Zhou, L.~Li, M.~Wang, and H.~Li, ``Motion-aware 3d gaussian splatting for efficient dynamic scene reconstruction,'' in \emph{European Conference on Computer Vision}, 2024.

\bibitem{lin2024gaussian}
Y.~Lin, Z.~Dai, S.~Zhu, and Y.~Yao, ``Gaussian-flow: 4d reconstruction with dynamic 3d gaussian particle,'' in \emph{Proceedings of the IEEE/CVF Conference on Computer Vision and Pattern Recognition}, 2024, pp. 21\,136--21\,145.

\bibitem{icml2024-sp-gs}
G.~Z. Diwen~Wan, Ruijie~Lu, ``Superpoint gaussian splatting for real-time high-fidelity dynamic scene reconstruction,'' in \emph{Forty-first International Conference on Machine Learning}, 2024.

\bibitem{li2022neural}
T.~Li, M.~Slavcheva, M.~Zollhoefer, S.~Green, C.~Lassner, C.~Kim, T.~Schmidt, S.~Lovegrove, M.~Goesele, R.~Newcombe \emph{et~al.}, ``Neural 3d video synthesis from multi-view video,'' in \emph{Proceedings of the IEEE/CVF Conference on Computer Vision and Pattern Recognition}, 2022, pp. 5521--5531.

\bibitem{zhang2018unreasonable}
R.~Zhang, P.~Isola, A.~A. Efros, E.~Shechtman, and O.~Wang, ``The unreasonable effectiveness of deep features as a perceptual metric,'' in \emph{Proceedings of the IEEE conference on computer vision and pattern recognition}, 2018, pp. 586--595.

\bibitem{li2022streaming}
L.~Li, Z.~Shen, zhongshu wang, L.~Shen, and P.~Tan, ``Streaming radiance fields for 3d video synthesis,'' in \emph{Advances in Neural Information Processing Systems}, A.~H. Oh, A.~Agarwal, D.~Belgrave, and K.~Cho, Eds., 2022.

\bibitem{attal2023hyperreel}
B.~Attal, J.-B. Huang, C.~Richardt, M.~Zollhoefer, J.~Kopf, M.~O'Toole, and C.~Kim, ``{HyperReel}: High-fidelity {6-DoF} video with ray-conditioned sampling,'' in \emph{Conference on Computer Vision and Pattern Recognition (CVPR)}, 2023.

\bibitem{song2023nerfplayer}
L.~Song, A.~Chen, Z.~Li, Z.~Chen, L.~Chen, J.~Yuan, Y.~Xu, and A.~Geiger, ``Nerfplayer: A streamable dynamic scene representation with decomposed neural radiance fields,'' \emph{IEEE Transactions on Visualization and Computer Graphics}, 2023.

\bibitem{duan:2024:4drotorgs}
Y.~Duan, F.~Wei, Q.~Dai, Y.~He, W.~Chen, and B.~Chen, ``4d-rotor gaussian splatting: Towards efficient novel view synthesis for dynamic scenes,'' in \emph{Proc. SIGGRAPH}, 2024.

\bibitem{Wang2023ICCV}
F.~Wang, S.~Tan, X.~Li, Z.~Tian, Y.~Song, and H.~Liu, ``Mixed neural voxels for fast multi-view video synthesis,'' in \emph{Proceedings of the IEEE/CVF International Conference on Computer Vision (ICCV)}, October 2023, pp. 19\,706--19\,716.

\bibitem{yangreal}
Z.~Yang, H.~Yang, Z.~Pan, and L.~Zhang, ``Real-time photorealistic dynamic scene representation and rendering with 4d gaussian splatting,'' in \emph{The Twelfth International Conference on Learning Representations}.

\bibitem{INSTA:CVPR2023}
\emph{Instant Volumetric Head Avatars}, 2023.

\bibitem{xiang2024flashavatar}
J.~Xiang, X.~Gao, Y.~Guo, and J.~Zhang, ``Flashavatar: High-fidelity head avatar with efficient gaussian embedding,'' in \emph{The IEEE Conference on Computer Vision and Pattern Recognition (CVPR)}, 2024.

\bibitem{ma2024gaussianblendshapes}
S.~Ma, Y.~Weng, T.~Shao, and K.~Zhou, ``3d gaussian blendshapes for head avatar animation,'' in \emph{ACM SIGGRAPH Conference Proceedings, Denver, CO, United States, July 28 - August 1, 2024}, 2024.

\end{thebibliography}

\end{document}